\documentclass[conference]{IEEEtran}
\IEEEoverridecommandlockouts
\usepackage{cite}
\usepackage{amsmath,amssymb,amsfonts}
\usepackage{algorithmic}
\usepackage{graphicx}
\usepackage{textcomp}
\usepackage{xcolor}
\def\BibTeX{{\rm B\kern-.05em{\sc i\kern-.025em b}\kern-.08em
    T\kern-.1667em\lower.7ex\hbox{E}\kern-.125emX}}

\newcommand \ignore[1]{}
\newcommand*{\affaddr}[1]{#1} 
\newcommand*{\affmark}[1][*]{\textsuperscript{#1}}
\newcommand*{\email}[1]{\texttt{#1}}
\usepackage{url}
\usepackage{colortbl}

\usepackage{caption}
\usepackage{subcaption}

\makeatletter
\def\ps@IEEEtitlepagestyle{%
  \def\@oddfoot{\mycopyrightnotice}%
  \def\@oddhead{\hbox{}\@IEEEheaderstyle\leftmark\hfil\thepage}\relax
  \def\@evenhead{\@IEEEheaderstyle\thepage\hfil\leftmark\hbox{}}\relax
  \def\@evenfoot{}%
}
\def\mycopyrightnotice{%
  \begin{minipage}{\textwidth}
  \scriptsize
  ~\copyright~2024 IEEE.  Personal use of this material is permitted.  Permission from IEEE must be obtained for all other uses, in any current or future media, including reprinting/republishing this material for advertising or promotional purposes, creating new collective works, for resale or redistribution to servers or lists, or reuse of any copyrighted component of this work in other works.
  \end{minipage}
}
\makeatother

\begin{document}

\title{
Mitigating Sex Bias in Audio Data-driven COPD and COVID-19 Breathing Pattern Detection Models \\
}

\author{Rachel Pfeifer\affmark[1], Sudip Vhaduri\affmark[2], James Eric Dietz\affmark[2]\\
\affaddr{\affmark[1]Computer Science Department,
\affmark[2]Computer and Information Technology Department, \\
Purdue University, IN 47907}\\
\email{\{\affmark[1]rjpfeife,\affmark[2]svhaduri,\affmark[2]jedietz}\}@purdue.edu}

\maketitle

\begin{abstract}
In the healthcare industry, researchers have been developing machine learning models to automate diagnosing patients with respiratory illnesses based on their breathing patterns. However, these models do not consider the demographic biases, particularly sex bias, that often occur when models are trained with a skewed patient dataset. Hence, it is essential in such an important industry to reduce this bias so that models can make fair diagnoses. 
In this work, we examine the bias in models used to detect breathing patterns of two major respiratory diseases, i.e., chronic obstructive pulmonary disease (COPD) and COVID-19. 
Using decision tree models trained with audio recordings of breathing patterns obtained from two open-source datasets consisting of 29 COPD and 680 COVID-19-positive patients, we analyze the effect of sex bias on the models. With a threshold optimizer and two constraints (demographic parity and equalized odds) to mitigate the bias, we witness 81.43\% (demographic parity difference) and 71.81\% (equalized odds difference) improvements. These findings are statistically significant.
\end{abstract}


\section{Introduction}\label{introduction}

\subsection{Motivation}
Each year, millions of people are affected by respiratory diseases such as COPD, in addition to coronavirus-caused diseases such as COVID-19. Recent studies show that in 2020, over 480 million people worldwide had COPD, and according to the World Health Organization (WHO), COPD is the third leading cause of death globally \cite{copdstats}. COVID-19 has also impacted the respiratory health of many individuals, starting in 2019 with the global pandemic. The WHO estimates that in 2020 and 2021 alone, 14.9 million people died due to COVID-19. Researchers note that COVID-19 has decreased global life expectancy by two years, signifying the severity of this illness' effect \cite{covidstats,covidlifeexpectancy}.

Hence, it is crucial to develop machine learning models that can diagnose prevalent respiratory illnesses such as COPD or COVID-19 using patients' breathing patterns captured in audio recordings via smartphones or other devices. However, there is a significant amount of sex bias in the healthcare industry, especially in respiratory health. Studies reveal that respiratory illnesses are more likely to be misdiagnosed or underdiagnosed in female patients than in male patients, with female patients often having the seriousness of their symptoms dismissed or not being given a referral to a specialist \cite{misdiagnosing}. This bias can bring into question the effectiveness of the model. Therefore, when developing machine learning models for diagnosing respiratory illnesses, it is necessary to incorporate the demographic bias, specifically sex bias, that exists in the healthcare industry.

\subsection{Related Work}
Studies have shown that female patients with respiratory illnesses, including COPD and COVID-19, were approximately 16\% more likely not to take breathing tests \cite{breathingtests}; these breathing tests are used by healthcare professionals to aid in the diagnosis of respiratory illnesses. Another study discusses the prevalence of underdiagnosing COPD. Researchers in this study find that underdiagnosing COPD is 30\% more common in female patients than male patients, with female patients reporting that physicians spend insufficient time with them \cite{insufficienttime}. Since respiratory illnesses are significantly less likely to be diagnosed in female patients compared to male patients, it is not surprising that the number of women who are either hospitalized or dying from these respiratory illnesses is greater than the number of men who experience this \cite{genderandcopd}.

To ease the diagnosis of widespread respiratory illnesses, researchers have recently started studying and developing deep-learning models using audio recordings  \cite{deeplearningaudiodata}. 
Some studies focus on specifically examining various types of audio data from patients, such as coughing, normal breathing, deep breathing, and voice recordings \cite{vhaduri2023environment,vhaduri2023transfer}. While some efforts aim to identify potential biases in medical applications of artificial intelligence (AI) in general~\cite{aibias}, current research does not address biases, specifically sex bias, in models developed with audio recordings to detect respiratory disease patterns and ways to mitigate them when diagnosing COPD and COVID-19.

\subsection{Contributions}
The contributions of our work involve analyzing how machine learning models developed with audio recordings of breathing patterns of patients infected with either COPD or COVID-19 are affected by demographic bias. 
We focus on sex bias since female patients are significantly underrepresented in respiratory health compared to male patients. By applying bias mitigation algorithms, we substantially mitigate the demographic bias among the sexes; we use a threshold optimizer with constraints of demographic parity and equalized odds as the bias mitigation algorithms.

\section{Methodology}\label{methods}

We discuss our methodology in this section to reduce bias in COPD and COVID-19 detection models that utilize breathing audio data. Before presenting this methodology and corresponding datasets with pre-processing steps, we introduce key terms.

\subsection{Preliminaries}\label{preliminaries}

\subsubsection{Fairness and Bias}
Per the FairLearn library~\cite{fairlearn}, we define fairness as the unfair impact or harm a model has on a group of individuals. Bias is the type of unfair impact that the group experiences. In this work, we explore the demographic bias in COPD and COVID-19 detection models, focusing specifically on the sex bias. Sex bias is the unfair behavior a model causes based on sex differences.

\subsubsection{Fairness Metrics}\label{fairMetrics}
To evaluate how our bias mitigation algorithms are performing, we delve into multiple metrics. We define the metrics we utilize below.

\begin{itemize}
    \item The selection rate is described as the proportion of data points classified as a positive (correct) result. These implications make it essential to increase the selection rate while decreasing the bias.
    \item The demographic parity ratio is defined as a comparison of the smallest and largest proportions of data points classified as a positive (correct) result. In other words, it is a comparison of the smallest and largest group selection rates. When the demographic parity ratio is 1, every group in a dataset has the same selection rate. It is important to increase this metric when reducing the bias.
    \item The demographic parity difference refers to the gap between the smallest and largest proportions of data points classified as a positive (correct) result. Simply put, it is the gap between the smallest and largest group selection rates. We aim to decrease the demographic parity difference as we decrease the bias.
    \item The false negative rate is a measure of how frequently models classify positive results incorrectly. Since a false negative rate is a type II error in statistics and involves failing to reject a null hypothesis that should be rejected, we seek to decrease the value of this metric.
    \item The equalized odds ratio is described as the smaller of the true positive rate and false positive rate ratios. An equalized odds ratio with a value of 1 indicates that the false negative rate, true positive rate, false positive rate, and true negative rate are all equal in a dataset. We aim to increase this metric when reducing the bias. 
    \item The equalized odds difference represents the larger of the true positive rate and false positive rate differences. While the true positive rate and false positive rate differences are like the true positive rate and false positive rate previously mentioned, we calculate the difference in these rates with the equalized odds difference. We look to decrease this metric as we mitigate the bias.
\end{itemize}

\subsubsection{Sensitive Features and Constraints}
We define sensitive features as the attributes that make up a person, such as sex or age. A constraint is the criteria that affect the sensitive feature by applying it to the bias mitigation algorithms. These algorithms use the same metrics for fairness on the sensitive feature by following the constraints; we use constraints of demographic parity and equalized odds in our bias mitigation algorithms. 

\subsubsection{Percentage Improvement}
We use percentage improvement to determine how our bias mitigation algorithms perform. To calculate the percentage improvement of a metric, we first find the absolute value of the difference between the value of the metric before and after mitigation. We then divide the absolute difference by the value of the metric before mitigation and multiply by 100.  

\begin{figure*}
\centering
\subfloat[Selection rate]{\includegraphics[width=.31\linewidth]{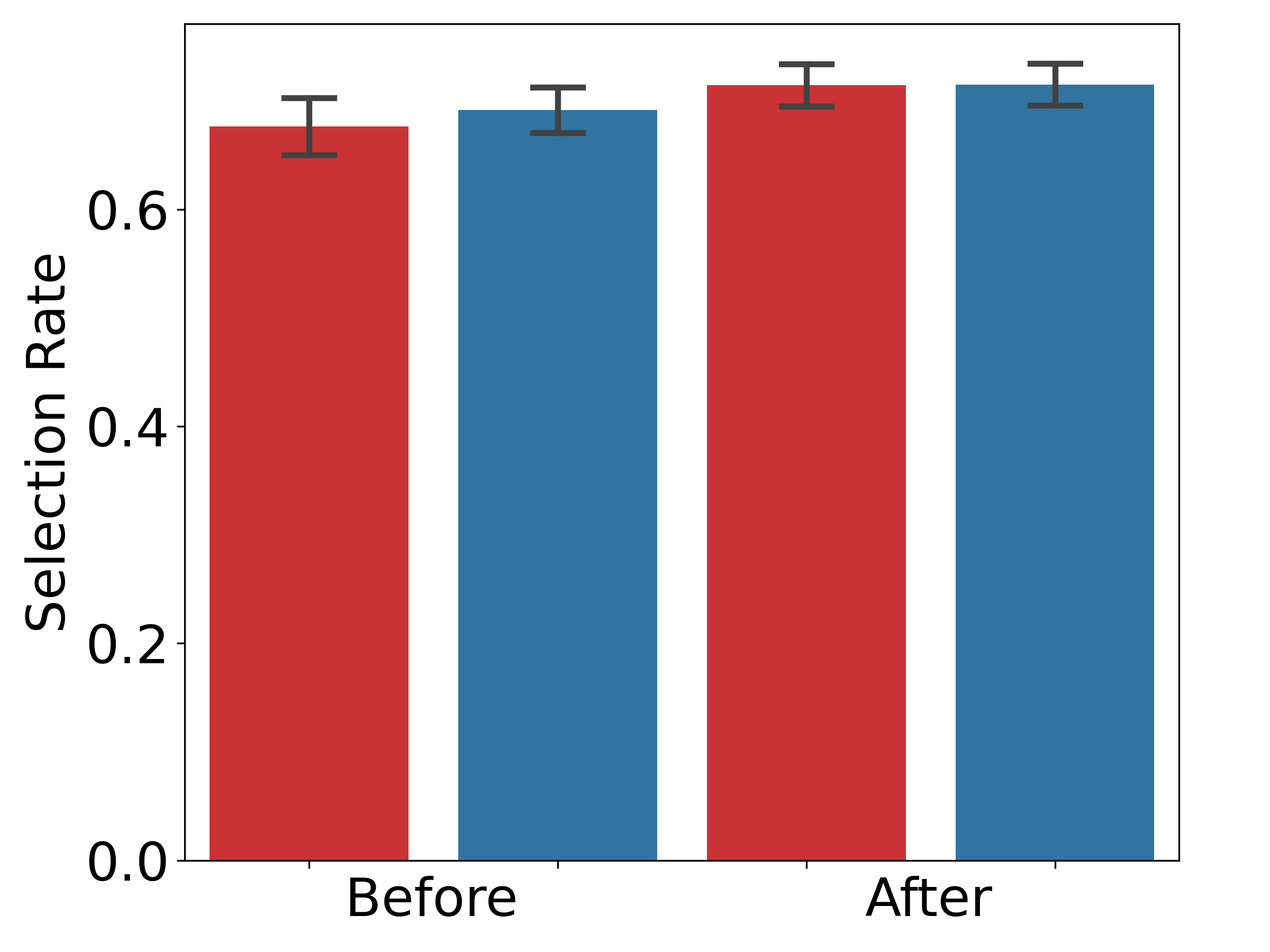}\label{selectionrate}}
\hfill
\subfloat[Demographic parity ratio]{\includegraphics[width=.31\linewidth]{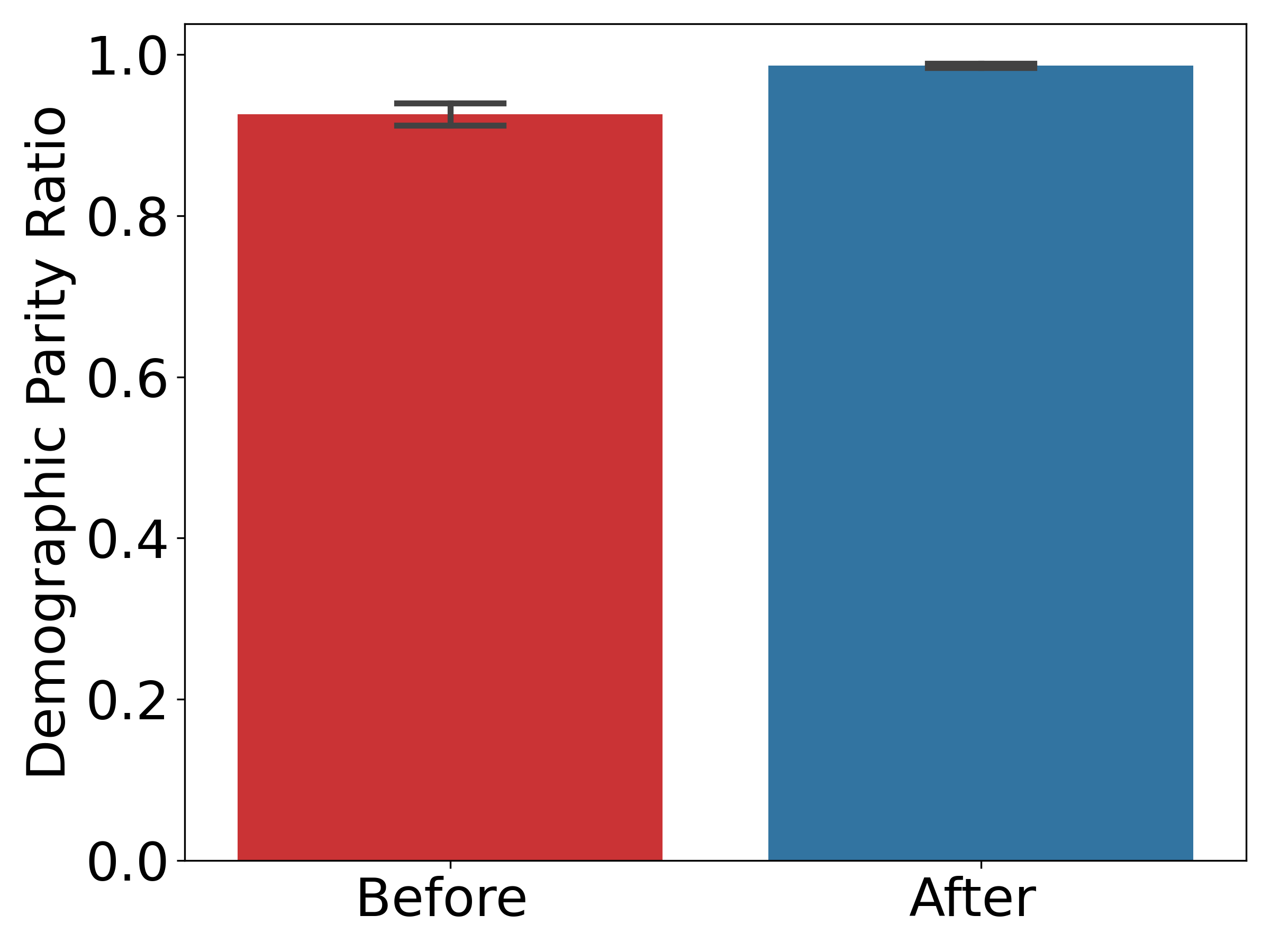}\label{demographicparityratio}}
\hfill
\subfloat[Demographic parity difference]{\includegraphics[width=.31\linewidth]{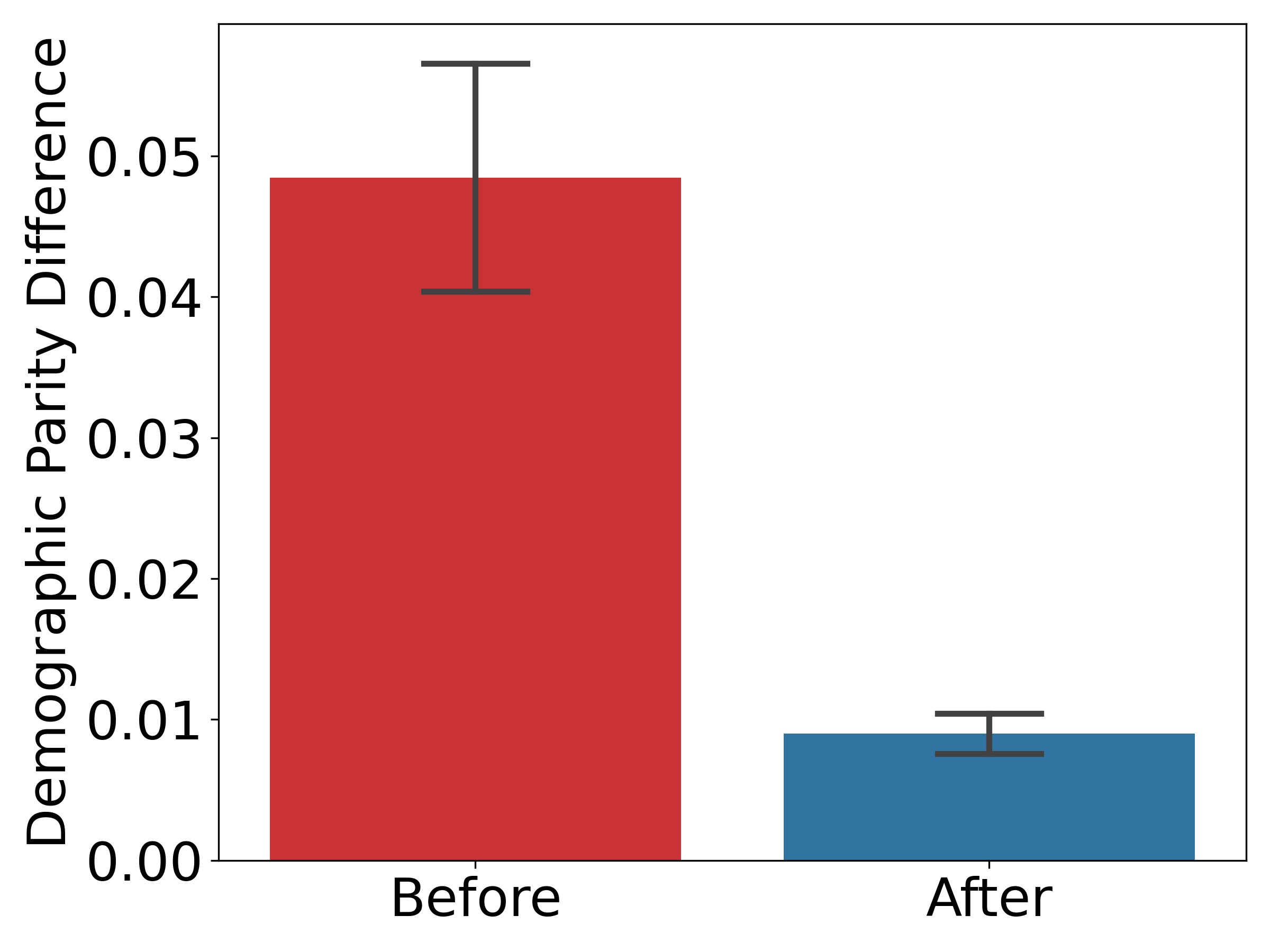}\label{demographicparitydifference}}
\caption
{Bar graph with error bars representing findings of the demographic parity analysis}
\label{demographicparity}
\end{figure*}

\subsection{Datasets}\label{dataCollect}
In this work, we consider two open-source datasets that contain audio recordings of breathing sounds from patients with two respiratory illnesses, i.e., COPD~\cite{copd} and COVID-19~\cite{covid}. There are 29 COPD patients (21 male and 8 female) and 680 COVID-19-positive patients (413 male and 267 female). While the COPD recordings contain normal, deep, and heavy breaths, the COVID-19 dataset contains deep and shallow breaths. We utilize the deep breathing recordings available in the datasets. In addition to audio recordings, each dataset contains patients' sex and age.

\subsection{Pre-processing and Feature Engineering}\label{preProc}

We find that the COVID-19 audio recordings are relatively shorter than the COPD recordings, and we have way more patients and audio recordings from the COVID-19 dataset. Therefore, we first calculate the average length of the COVID-19 recordings, and we find the average length is $14.37 \pm 6.17$ seconds. Next, we choose this $14$ seconds as a threshold to pick the audio recordings, i.e., patients from the COVID-19 dataset. We end up with 319 patients. Since the COPD recordings are longer than the threshold, we ended up with all 29 patients. Next, from each audio recording, we select the first 14 seconds and segment it into 2-second long seven smaller recordings. Using another Python library called Librosa, we obtain 40 MFCC (Mel-Frequency Cepstral Coefficient)~\cite{MFCCs} features for each patient's seven segmented deep breathing audio recordings. We also filter out patients with MFCC features equal to 0. 

\subsection{Data Modeling}
In this work, we use binary decision tree classification learners to distinguish COPD and COVID-19 breathing patterns. We develop two decision tree models: one for demographic parity and one for equalized odds. 
The feature set we use for our models includes sex (i.e., the sensitive feature), age, and 40 MFCC features. Since we end up with 203 COPD instances (7 from one of the 29 patients), for class balancing, we randomly select 203 COVID-19 instances (142 instances from males and 61 from females).

\subsection{Hyper-parameter Optimization}
To determine the optimal values of our decision tree's hyper-parameters, we use grid search cross-validation (CV). 
We optimize the $criterion$, $min\_samples\_leaf$, and $min\_samples\_split$. The $criterion$ has possible selections of ``gini'' or ``entropy.'' The values for $min\_samples\_leaf$ and $min\_samples\_split$ are both $\{2,3,4,5\}$. To establish a bias-variance trade-off, we find $CV = 5$ to be the best choice in our case. Here, our optimal hyper-parameters are  ``gini'' for the criterion, $3$ for $min\_samples\_leaf$, and $4$ for $min\_samples\_split$.

\subsection{Bias Mitigation}
To mitigate the sex bias in our dataset and models, we use a threshold optimizer with constraints of demographic parity and equalized odds from a Python library called FairLearn~\cite{fairlearn}. We define a threshold optimizer as a post-processing method that involves using a different threshold for each of the estimator's groups. The group we utilize is sex, and we have a decision tree as our estimator. We evaluate the threshold optimizer on three different metrics for each constraint, i.e., demographic parity and equalized odds. For both before and after mitigation, we calculate each of the metric values 30 times. Our work is accessible in GitHub~\cite{breathinggithub}

\begin{figure*}
\centering
\subfloat[False negative rate]{\includegraphics[width=.31\linewidth]{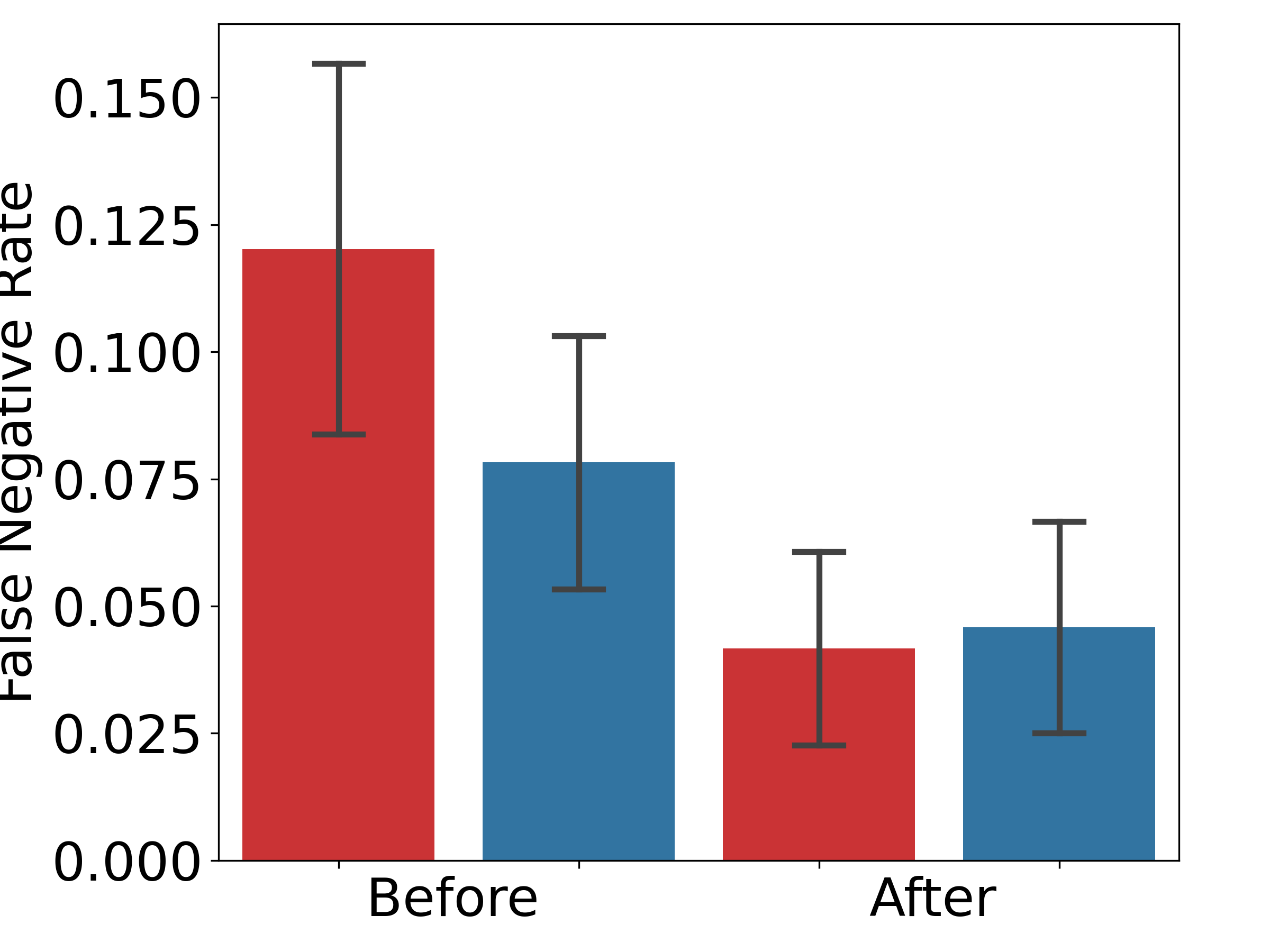}\label{falsenegativerate}}
\hfill
\subfloat[Equalized odds ratio]{\includegraphics[width=.31\linewidth]{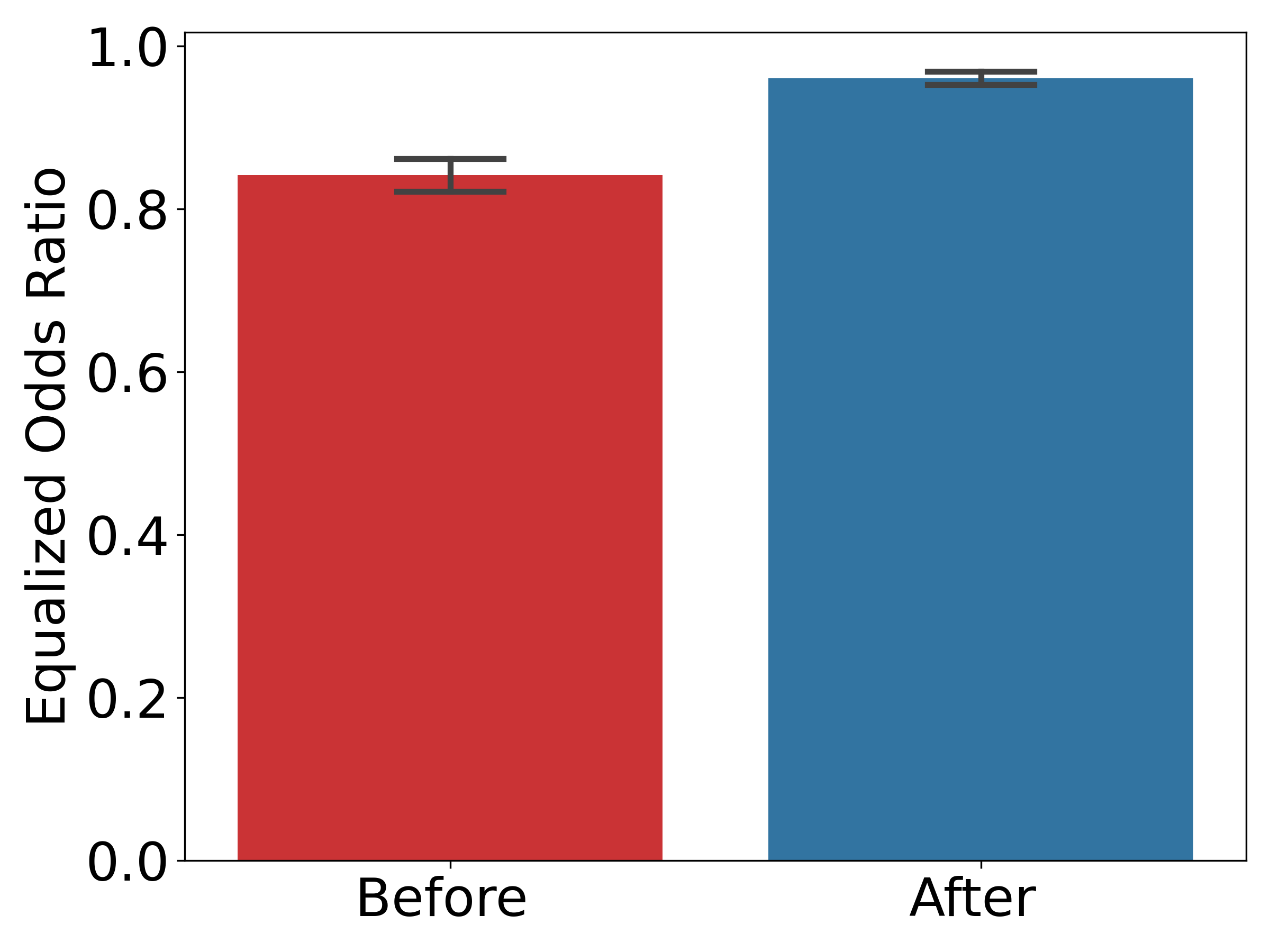}\label{equalizedoddsratio}}
\hfill
\subfloat[Equalized odds difference]{\includegraphics[width=.31\linewidth]{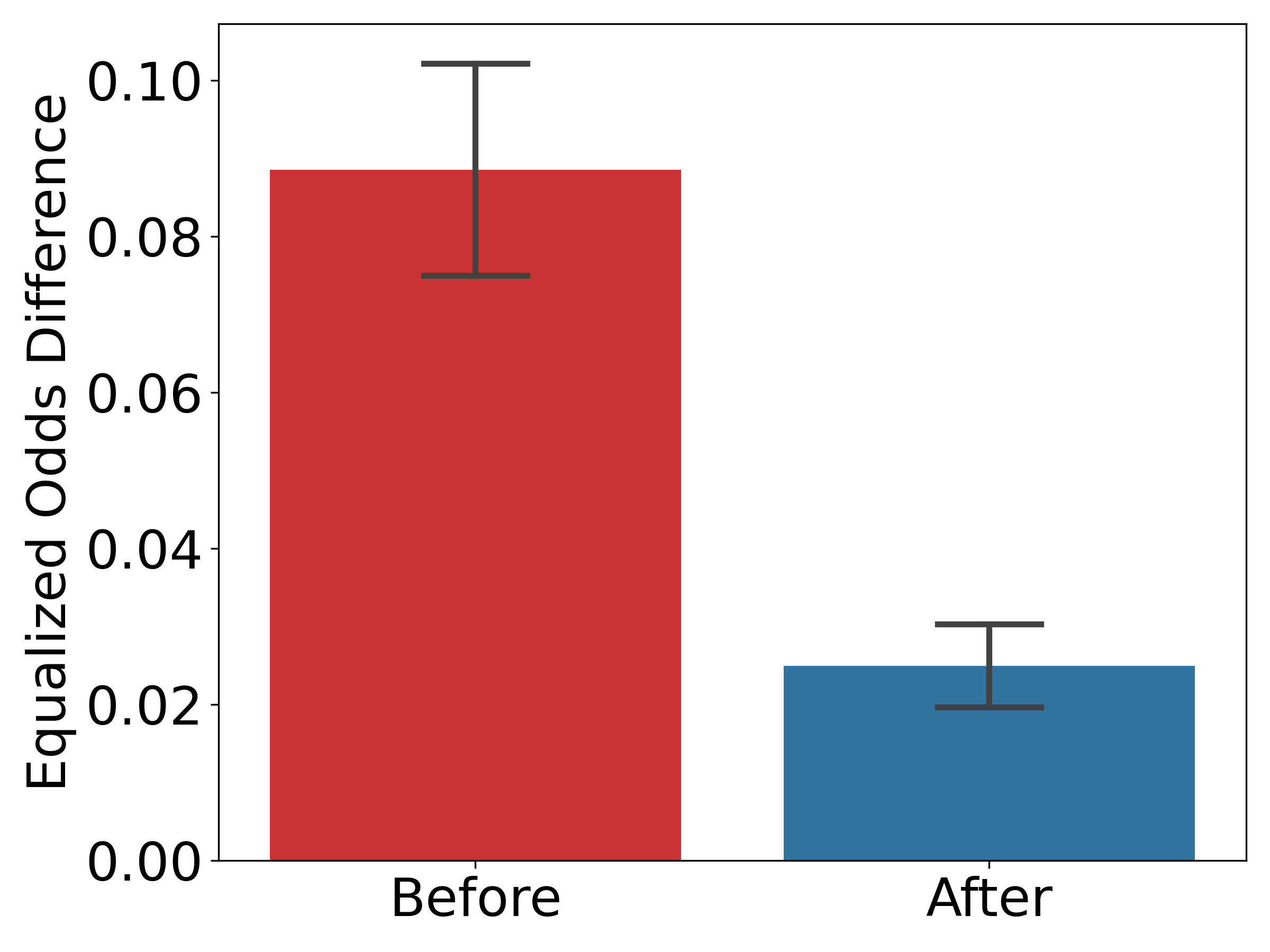}\label{equalizedoddsdifference}}
\caption{Bar graph with error bars representing findings of the equalized odds analysis}
\label{equalizedodds}
\end{figure*}

\section{Analysis}\label{analysis}

Using graphical representation and statistical tests, we analyze the results of mitigating the sex bias. We utilize percentage improvement to quantify the changes in metric values after applying the mitigation algorithms. In Section~\ref{dpAnalysis}, we assess the evaluation of the demographic parity constraint, and in Section~\ref{eoAnalysis}, we evaluate the equalized odds constraint.

\subsection{Evaluation of Demographic Parity}\label{dpAnalysis}
We use the metrics of selection rate, demographic parity ratio, and demographic parity difference to analyze how bias mitigation affects the demographic parity constraint.

\subsubsection{Comparison of Selection Rate Metric} 
In Figure~\ref{selectionrate}, we find that before mitigation, the female (red bars) selection rate is 67.64\%, and the male (blue bars) selection rate is 69.15\%. For both sexes, the selection rates increase after mitigation, with the female selection rate rising to 71.42\% and the male selection rate growing to 71.50\%. These rates are more alike following the mitigation of the bias, and we can state that the mitigation algorithm increases the average selection rate, as seen in the increase from 68.39\% to 71.46\%.

\subsubsection{Comparison of Demographic Parity Ratio Metric}
In Figure~\ref{demographicparityratio}, the demographic parity ratio is 92.62\%, and after mitigation, it is 98.64\%. Thus, reducing the bias leads to a 6.50\% improvement in the demographic parity ratio.

\subsubsection{Comparison of Demographic Parity Difference Metric}
Figure~\ref{demographicparitydifference} depicts the demographic parity difference as 4.85\% before mitigation. This value decreases to 0.90\% after mitigation, resulting in an 81.43\% improvement.

\subsubsection{Two-sample t-test}
To see if the demographic parity difference is statistically significant, we utilize \emph{Welch's} version of the {\em t-test}. We define \emph{$H_0: \mu_b = \mu_a$} as our null hypothesis. The mean demographic parity difference before mitigating the bias is \emph{$\mu_b$}, and the mean demographic parity difference after mitigating the bias is \emph{$\mu_a$}. The \emph{Welch's Two-sample t-test} yields \emph{$t(30.80) = 4.81$}, with \emph{p-value = $3.75e^{-5}$}. Since \emph{$3.75e^{-5} < 0.05$}, we reject our null hypothesis. Therefore, the difference between the average demographic parity differences prior to and following mitigation is significant.

\subsection{Evaluation of Equalized Odds}\label{eoAnalysis}
We explore the false negative rate, equalized odds ratio, and equalized odds difference metrics to determine how bias mitigation impacts the equalized odds constraint.

\subsubsection{Comparison of False Negative Rate Metric}
In Figure~\ref{falsenegativerate}, we observe that the female (red bars) and male (blue bars) false negative rates before mitigation are 12.02\% and 7.82\%, respectively. These rates decrease following mitigation, with the female false negative rate falling to 4.17\% and the male false negative rate dropping to 4.58\%. These rates are more proportionate after mitigation, and we can assert that the mitigation algorithm causes the average false negative rate to decrease since the average rate falls from 9.92\% to 4.37\%.

\subsubsection{Comparison of Equalized Odds Ratio Metric}
In Figure~\ref{equalizedoddsratio}, we find that the equalized odds ratio increases from 84.17\% before mitigation to 96.06\% after mitigation, resulting in a 14.13\% improvement.

\subsubsection{Comparison of Equalized Odds Difference Metric}
In Figure~\ref{equalizedoddsdifference}, we see an equalized odds difference of 8.86\% before mitigation, decreasing to 2.50\% after mitigation. This is a 71.81\% improvement.

\subsubsection{Two-sample t-test}
We employ \emph{Welch's} version of the {\em t-test} to determine if the equalized odds difference is statistically significant. We assert that \emph{$H_0: \mu_b = \mu_a$} is the null hypothesis. The mean equalized odds difference prior to mitigating the bias is \emph{$\mu_b$}, and the mean equalized odds difference following mitigating the bias is \emph{$\mu_a$}. The \emph{Welch's Two-sample t-test} results in \emph{$t(37.67) = 4.36$} and \emph{p-value = $9.72e^{-5}$}. Since \emph{$9.72e^{-5} < 0.05$}, we reject the null hypothesis. Thus, we can conclude that the equalized odds difference results are significant.

\section{Discussion}\label{discussion}

To our best knowledge, we are the first to present work that aims to mitigate bias in COPD and COVID-19 breathing pattern detection models. Through a detailed analysis of two open-source datasets, we demonstrate that our mitigation algorithms, developed with a threshold optimizer with demographic parity and equalized odds constraints, significantly reduce the sex bias across male and female groups.

There are some limitations to our work. First, our dataset consists of patients from two open-source datasets (as mentioned previously), one of which has a small number of patients. To increase the sample size of our datasets, we take seven different segments of each patient's deep breathing audio recordings, but we develop models with cross-validation and mutual exclusion. Therefore, our results are still indicative of bias reduction. A second limitation we observe is mitigating sex bias. Sensitive features like age can impact how a machine learning model performs. However, the age distribution of all patients in the used datasets is homogeneous.  

Future work pertains to pooling patients from larger respiratory datasets that contain a variety of sensitive features, including age and race, with a wider range of data distributions. Another direction is to evaluate how training a decision tree with different types of breathing recordings (e.g., normal or heavy breathing instead of or in addition to the deep breathing that we use in this work) impacts the model performance and the bias present. This work can further be extended to other domains, such as security via biometric authentication~\cite{vhaduri2024mwiotauth,vhaduri2023implicit}, where bias analysis did not get attention.

\bibliographystyle{IEEEbib}
\bibliography{main}

\end{document}